\documentclass[a4paper,11pt]{article}

\usepackage[english]{babel}
\usepackage[utf8]{inputenc}  
\usepackage[T1]{fontenc}
\usepackage{graphicx}
\usepackage{amsmath,amssymb,bm,arydshln}
\usepackage{multicol}
\usepackage{geometry}
\usepackage{natbib}
\usepackage{subcaption}
\usepackage{graphicx}
\graphicspath{{Figures/}}
\usepackage{cite}
\usepackage{booktabs}
\usepackage{booktabs}
\usepackage[font={scriptsize},labelfont=bf]{caption}
\usepackage[]{hyperref}
\hypersetup{pdftex,colorlinks=true,allcolors=black}
\usepackage{enumitem}
\usepackage{gensymb}
\usepackage{verbatim}
\usepackage{enumitem}
\usepackage[final]{pdfpages}

\newcommand\E{\mathbb{E}}

\DeclareMathOperator*{\argmax}{arg\,max}
\newcommand{\T}{^{\top}}

\newcommand{\x}{\boldsymbol{x}}

\newcommand{\z}{\boldsymbol{z}}
\newcommand{\s}{\boldsymbol{\Sigma}}
\newcommand{\m}{\boldsymbol{\mu}}
\newcommand{\h}{\boldsymbol{\theta}}
\newcommand\V{\mathbb{V}}

\makeatletter
\newcommand{\verbatimfont}[1]{\def\verbatim@font{#1}}%
\makeatother

\makeatletter
\newcommand\code{\bgroup\@makeother\_\@makeother\~\@makeother\$\@codex}
\def\@codex#1{{\normalfont\ttfamily\hyphenchar\font=-1 #1}\egroup}
\makeatother
\newcommand{\pkg}[1]{{\fontseries{b}\selectfont #1}}

\usepackage{microtype}
\RequirePackage[scaled=0.92]{helvet}
\RequirePackage{palatino,mathpazo}
\RequirePackage[scaled=1.02]{inconsolata}

\title{Handling missing data in model-based clustering}

\author{%
Alessio Serafini\textsuperscript{1},
Thomas Brendan Murphy\textsuperscript{2},
Luca Scrucca\textsuperscript{1}\\[1em]
\small\textsuperscript{1}  Department of Economics, Universit\`{a} degli Studi di Perugia, Italy\\
\small\textsuperscript{2} School of Mathematics and Statistics, University College Dublin, Ireland\\[1em]
}

\begin{document}

\maketitle

\begin{abstract}
Gaussian Mixture models (GMMs) are a powerful tool for clustering, classification and density estimation when clustering structures are embedded in the data. The presence of missing values can largely impact the GMMs estimation process, thus handling missing data turns out to be a crucial point in clustering, classification and density estimation. Several techniques have been developed to impute the missing values before model estimation. Among these, multiple imputation is a simple and useful general approach to handle missing data. 
In this paper we propose two different methods to fit Gaussian mixtures in the presence of missing data. 
Both methods use a variant of the Monte Carlo Expectation-Maximisation (MCEM) algorithm for data augmentation. Thus, multiple imputations are performed during the E-step, followed by the standard M-step for a given eigen-decomposed component-covariance matrix.
We show that the proposed methods outperform the multiple imputation approach, both in terms of clusters identification and density estimation.
\end{abstract}

\section{Introduction}
In many real applications, observations are subject to missing entries during the data collection process. Thus, handling these missing values is a crucial point in model estimation. This aspect is also fundamental in pattern recognition, where missing values can be informative, and an inaccurate treatment can lead to serious errors in classification and density estimation.
Before focusing on the methods available for dealing with missing data, it is necessary to specify the nature of the missing mechanism \citep{Rubin:1976}. 
This describes how the probability of a missing value is related to the data and, although in general it is unknown and unobservable, it is necessary to make some assumptions about the underlying missing mechanism.
Often the validity of the methods used in practical applications depend on whether or not these assumptions are fulfilled. 
Three different missing mechanisms are common in the literature: Missing Completely At Random (MCAR), Missing At Random (MAR), and Missing Not At Random (MNAR). For the technical details and definitions we refer to \citet{Little:Rubin:2002}. 
We note, however, that under the MCAR and MAR assumptions the missing mechanism is considered to be ignorable, i.e. model parameters are not affected by the distribution of the missing data. 

Different approaches for dealing with missing values can be found in the literature. 
In the listwise/casewise deletion approach observations with missing entries are removed, and models are estimated using only the full set of observed data \citep[Chap. 3]{Little:Rubin:2002}. 
This approach is simple and fast to implement, but the estimates may be biased due to the loss of information caused by the removal of part of the data. 
Another popular approach is to impute the missing observations, that is somehow fill in the missing data with a single value or a set of values. 
Single imputation methods are the mean imputation, which fill the missing values with the mean or the conditional mean \citep{Wilks:1932}, the regression imputation \citep{Buck:1960}, which replaces missing values with predicted scores from a regression model, and its stochastic version \citep[Sec. 4.3]{Little:Rubin:2002}, the hot-deck imputation \citep{Ford:1983}, which is a collection of techniques that impute the missing values with scores from similar observations. 
Multiple imputation refers to a set of procedures that fill the missing values with different plausible values, thus generating different complete datasets.
Then, each complete dataset is analysed and the results are combined to reflect the uncertainty due to the presence of missing values \citep{rubin1987multiple, schafer1997analysis}. 
Finally, model-based procedures assume a specific distribution for the observed data and draw inference on parameters based on the likelihood. A popular method for Maximum-Likelihood in missing-data problems is the Expectation-Maximisation (EM) algorithm \citep{dempster1977maximum}. 

Gaussian Mixture Models (GMMs) are a powerful tool for density estimation, clustering, and classification \citep{fraley2002model, mclachlan2000finite}. Parsimonious parametrisation of covariance matrices allows for a flexible class of models, each with its distinctive characteristics \citep{banfield1993model, celeux1995gaussian}. 
However, the resulting log-likelihood is difficult to maximise directly, even numerically (see McLachlan and Peel, 2000, Sect. 2.8.1), so GMMs are usually fitted by reformulating the mixture problem as an incomplete-data problem within the EM framework.
Since the missing observations may contain important information about the clustering structure of the data, \citet{ghahramani1995learning} extended the EM algorithm for GMMs with missing values. They derived a closed-form solution for the M-step, but only in the unconstrained case, that is for full within-cluster covariance matrices. However, no solutions are available for the parsimonious parametrisations of covariance matrices mentioned above.

In this paper we extend the work of \citet{ghahramani1995learning} by considering all the parsimonious covariance structures proposed in \citet{banfield1993model} and \citet{celeux1995gaussian}. 
In our proposal, we use a modified E-step based on augmented data, followed by maximisation of the complete-data log-likelihood using the standard M-step for GMMs. 
To this goal, we have developed two model-based approaches to handle missing data in GMMs. 
Assuming that the data are distributed as multivariate Gaussian within each cluster, Monte Carlo sampling allows to approximate the E-step using an augmented dataset with filled missing values. The complete-data log-likelihood of the augmented dataset can then be maximised using standard formulas for GMMs. The main advantage of this approach is that only the E-step is modified, whereas the M-step for each parsimonious parametrisation of the covariance matrices is not affected.

The paper is organised as follows. Section 2 contains a brief description of Gaussian mixtures and the standard EM algorithm. Section 3 introduces two approaches for handling missing data in Gaussian mixture models, with details on the corresponding Monte Carlo EM algorithms presented in Section 4. In Section 5 the performance of the proposed algorithms are evaluated on both simulated and real datasets in terms of both clustering and density estimation. The final section provides some concluding remarks.

\section{Background}
\subsection{Gaussian Mixture Models}
\label{sec:gmm}

Gaussian Mixture Models (GMMs) are a popular parametric model for cluster analysis and density estimation. This class of models has been widely used for different problems in several fields for its flexibility and interpretability. Extensive reviews of mixture models can be found in \citet{mclachlan2000finite} and in \citet{fraley2002model} for the applications of GMMs to cluster analysis, discriminant analysis, and density estimation.

Let $\x = (\x_1, \x_2, \ldots, \x_N)$ be a random sample of $N$ observations with $\x_i \in \mathbb{R}^d$. 
The observed data are assumed to arise from a Gaussian mixture distribution if the density can be written as a convex combination of multivariate Gaussian distributions such that: 
\begin{equation*} 
\label{general mixture}
f( \x_i; \h) = \sum^{G}_{g=1} \pi_{g} \phi( \x_i ; \m_{g},\s_{g}),
\end{equation*}
where $G$ is the number of mixture components, $\phi( \x_i | \m_{g},\s_{g})$ is the underlying density function of the $g$th component, a multivariate Gaussian distribution with mean $\m_{g}$ and covariance matrix $\s_{g}$, $(\pi_1,\pi_2, \ldots, \pi_G)$ are the mixing weights, such that $\pi_g > 0$ and $\sum_{g=1}^{G} \pi_g = 1$, and $\h = \{\pi_1, \ldots, \pi_{G-1}, \m_{1}, \ldots, \m_{g}, \s_{1}, \ldots, \s_{g} \}$ is the set of unknown parameters of the mixture model.

The Gaussian assumption implies ellipsoidal clusters, each centred at the mean vector $\m_g$ and with covariance matrices $\s_g$. The latter control the geometrical features of the ellipsoids, such as the orientation, the volume and the shape.
Parsimonious parameterisation of covariance matrices for GMMs can be achieved through the eigen-decomposition $\s_g = \lambda_g\boldsymbol{D}_g\boldsymbol{A}_g\boldsymbol{D}\T_g$, where $\lambda_g$  is a scalar controlling the volume, $\boldsymbol{A}_g$ is a diagonal matrix controlling the shape, and $\boldsymbol{D}_g$ is an orthogonal matrix controlling the orientation of the ellipsoids. This parameterisation generates a broad class of models with specific geometric properties \citep{banfield1993model, celeux1995gaussian}.

Maximum likelihood estimation of the unknown parameters of a mixture model can in principle be obtained by maximising the log-likelihood:
\begin{equation*}
\ell(\h) = \sum_{i=1}^N \log \sum_{g=1}^G \pi_g \phi( \x_i ; \m_{g},\s_{g}).
\end{equation*}
However, the objective function above is hard to maximise directly, even numerically \citep[see][Sect. 2.8.1]{mclachlan2000finite}. Consequently, the standard algorithm employed for parameters estimation in mixture models is the Expectation-Maximisation (EM) algorithm, which is briefly reviewed in the following subsection.

\subsection{EM algorithm for mixture models}

\citet{dempster1977maximum} introduced an iterative procedure to estimate the parameters of a mixture model by maximising the log-likelihood by alternating two different steps. The Expectation step (E-step) computes the expected value of complete-data log-likelihood, and the Maximisation step (M-step) maximises the expected value previously computed with respect to the parameters of the mixture model. The algorithm guarantees a non-decreasing log-likelihood and, under fairly general conditions, it converges to at least a local maximum \citep{mclachlan2008algorithm}.

Suppose there exists an unobservable process $\z = (\z_1, \ldots, \z_N)$, where each $\z_i = (z_{i1}, \ldots, z_{iG})\T$ is a latent variable specifying the component-label, i.e. $z_{ig} = 1$ if the $i$-th observation belongs to component $g$ and $0$ otherwise. 

Let $f(\x,\z| \h)$ be the joint density of the complete-data vector, formed by the observed variable $\x$ and the latent variable $\z$. Then, the complete-data log-likelihood can be written as
\begin{equation}
\ell_c(\h) = \sum_{i=1}^{N} \log \{ f(\z_{i} ; \h)f(\x_{i} | \z_{i} ; \h) \}.
\label{logL}
\end{equation}
The E-step computes the expected value of the complete-data log-likelihood in \eqref{logL} with respect to the latent variable and using the current fit for $\h$, the so-called Q-function:
\begin{equation}
Q(\h | \h^{t}) = \E [ \ell_c(\h) ] 
= \E \left [ \sum_{i=1}^{N} \log f(\z_{i} ; \h)f(\x_{i} | \z_{i} ; \h) \right ].
\label{estep::gen}
\end{equation}
The M-step updates the estimate of $\h$ by maximising the Q-function in \eqref{estep::gen}, such that: 
\begin{equation*}
\h^{t+1} = \argmax_{\h} Q(\h | \h^{t}).
\end{equation*}

In the GMMs case, the complete-data log-likelihood is
\begin{equation}
\ell_c(\h) = \sum_{i=1}^{N} \sum_{g=1}^{G} z_{ig} \{ \log \pi_{g} + \log\phi(\x_i ; \m_g,\s_g) \}.
\label{loglikComp::mixture}
\end{equation}
At iteration $t$ of the EM algorithm, the Q-function is 
\begin{align*}
Q(\h | \h^{t}) & =  \sum_{i=1}^{N} \sum_{g=1}^{G} \E \left [ z_{ig} \{\log \pi_{g}^t + \log\phi(\x_i ; \m_g^t,\s_g^t)\} \right ],
\label{Q::mixture}
\end{align*}
so the E-step computes the expected value of $z_{ig}$ as 
\begin{equation*}
\hat{z}_{ig}^t = \E[z_{ig}|\x_i] = \dfrac{\pi_g^{t-1} \phi(\x_i ; \m_g^{t-1},\s_g^{t-1})}{\sum_{k=1}^G  \pi_k^{t-1} \phi(\x_i ; \m_k^{t-1},\s_k^{t-1})}.
\end{equation*}
The parameters of the mixture components are updated in the M-step by maximising 
\begin{equation*}
Q(\h | \h^{t}) = \sum_{i=1}^{N} \sum_{g=1}^{G} \hat{z}_{ig}^t \{ \log \pi_{g} + \log \phi(\x_i ; \m_g,\s_g) \},
\end{equation*}
with respect to the parameters $(\pi_g,\m_g,\s_g)$ for $g=1,\ldots,G$. The procedure is repeated until convergence, and the estimated parameters $(\hat{\pi}_g, \hat{\m}_g, \hat{\s}_g)$ for $g=1,\ldots,G$ are returned. 
\citet{celeux1995gaussian} provide details of the M-step for different covariance parameterisations, some of which have closed-form solutions, while others required numerical procedures.

\section{Missing data in Gaussian mixture models}

\subsection{Preliminaries}

Using the previous notation, the single observation $\x_i$ is partitioned as $\x_i = [\x_i^{(o)}, \x_i^{(m)}]\T$, where $\x_i^{(o)}$ is the observed part, and $\x_i^{(m)}$ is the missing part of the data vector. 
Note that, because the missing pattern for each observation could be different, the subscript $i$ should be included on the superscripts $(o)$ and $(m)$, but for ease of presentation and reading we avoid that notation.
If the vector $\x_i$ is assumed to be Gaussian, that is $[\x^{(o)}_i, \x^{(m)}_i]\T \sim \mathcal{N}(\m,\s)$, with 
\begin{align*}
\m & = \begin{bmatrix}
\m_{o} \\
\m_{m} 
\end{bmatrix},
\intertext{and}
\s & = \begin{bmatrix}
\s_{oo} & \s_{om}\\
\s_{mo} & \s_{mm} 
\end{bmatrix},
\end{align*}
where $\s_{mo} = \s_{om}\T$, then, by the conditional property of the Gaussian distribution, we may write the conditional distribution of the missing part given the observed part of the data as
\begin{equation*}
\x^{(m)}_i | \x^{(o)}_i \sim \mathcal{N}(\m^{(m)},\s^{(m)}),
\end{equation*}
with
\begin{align*}
\m^{(m)} & = \m_{m} + \s_{mo} \s_{oo}^{-1} (\x_i^{(o)} - \m_{o}), \\
\intertext{and}
\s^{(m)} & = \s_{mm} - \s_{mo} \s_{oo}^{-1} \s_{om}.
\end{align*}
Thus, the conditional distribution of the missing part given the observed part follows a multivariate Gaussian distribution, with $\E[\x^{(m)}|\x^{(o)}] = \m^{(m)}$ and $\V[\x^{(m)}|\x^{(o)}] = \s^{(m)}$. 

In the GMMs context, the underlying density function of each component is assumed to be a multivariate Gaussian distribution, that is $\x_i|(z_{ig}=1) \sim \mathcal{N}(\m_g,\s_g)$. 
In the presence of missing values, we may write:
\begin{equation*}
\begin{bmatrix}
\x_i^{(o)} \\ 
\x_i^{(m)}
\end{bmatrix} \biggl| (z_{ig}=1) 
\sim \mathcal{N} \left( 
\begin{bmatrix}
\m_{o,g} \\ 
\m_{m,g}
\end{bmatrix},
\begin{bmatrix}
\s_{oo,g} & \s_{om,g}\\
\s_{mo,g} & \s_{mm,g}
\end{bmatrix}
\right),
\end{equation*}
then
\begin{equation*}
\x_i^{(o)}|(z_{ig}=1) \sim \mathcal{N} \left( \m_g^{(o)}, \s_g^{(o)} \right)
\end{equation*}
with 
$\m_g^{(o)} = \m_{o,g}$, $\s_g^{(o)} = \s_{oo,g}$, and 
\begin{equation*}
\x_i^{(m)}|(\x_i^{(o)},z_{ig}=1) \sim \mathcal{N} \left( \m_{g}^{(m)},\s_g^{(m)} \right),
\end{equation*}
with 
\begin{align*}
\m^{(m)}_g & = \m_{m,g} + \s_{mo,g} \s_{oo,g}^{-1} (\x_i^{(o)} - \m_{o,g}), \\ 
\intertext{and}
\s^{(m)}_g & = \s_{mm,g} - \s_{mo,g} \s_{oo,g}^{-1} \s_{om,g} .
\end{align*}
Furthermore, by the conditional property of the Gaussian distribution, the joint distribution of the observed part and the missing part given the group membership can be 
factorised as the product of two Gaussian distributions:
\begin{equation*} 
\phi_g(\x_{i}^{(o)}, \x_{i}^{(m)} ; \h_{g}) = \phi(\x^{(o)}_i ; \m^{(o)}_g,\s^{(o)}_g)\phi(\x^{(m)}_i | \x^{(o)}_i ; \m^{(m)}_g,\s^{(m)}_g).
\end{equation*} 

\citet{ghahramani1995learning} proposed an extension of the standard EM algorithm to deal with missing values \citep[see also][]{hunt2003mixture}. 
The complete-data log-likelihood in this case can be written as
\begin{equation*}
\ell_c(\h) =  \sum_{i=1}^N \sum_{g=1}^G z_{ig} 
\left\{ \log\pi_g + 
        \log\phi(\x^{(o)}_i ; \m^{(o)}_g,\s^{(o)}_g) + 
        \log\phi(\x^{(m)}_i|\x^{(o)}_i ;\m^{(m)}_g,\s^{(m)}_g) 
\right\}.
\end{equation*}
Thus, the conditional expectation in the E-step takes the following form:
\begin{multline}
Q(\h | \h^{t}) =  \sum_{i=1}^N \sum_{g=1}^G \E
\biggl[ z_{ig} \left\{
  \log\pi_g +
\right. \\ 
\left.
  \log\phi(\x^{(o)}_i ; \m^{(o)}_g,\s^{(o)}_g) + 
  \log\phi(\x^{(m)}_i|\x^{(o)}_i ; \m^{(m)}_g,\s^{(m)}_g) 
\right\} \biggl| \x^{(o)}_i \biggr]. 
\label{Qmiss}
\end{multline}
In this case there are two sources of missingness, one related to the unknown classification, and one connected with the missing part of the data. Therefore, additional expectations must be introduced to solve the E-step. By solving these expectations, it is possible to obtain a closed-form expression for the M-step under the assumption of unconstrained within-component covariance matrices \citep{ghahramani1995learning, hunt2003mixture, eirola2014mixture}. 
However, the flexible parameterisations of the covariance matrices described in Section~\ref{sec:gmm} have not been taken into account. 
For this reason, we aim at proposing methods to solve the $Q$-function in \eqref{Qmiss} for the general family of parsimonious GMMs.

\subsection{Proposed methods}

In this paper we propose two versions of a Missing Monte Carlo EM (MMCEM) approach. Both versions are based on the idea of using a Monte Carlo EM algorithm \citep[MCEM;][]{wei1990monte} to approximate the expected values required in the E-step.

Starting from equation~\eqref{Qmiss}, the first idea is to directly use MC approximations to solve the additional expectations generated from the presence of missing values. The expected complete-data log-likelihood can be approximated as follows:
\begin{multline}
Q(\h | \h^{t}) \approx \frac{1}{S}  \sum_{s=1}^S \sum_{i=1}^N \sum_{g=1}^G 
\hat{z}_{ig,s} 
\left\{
  \log\pi_g +
\right.\\
\left.
  \log\phi(\x^{(o)}_i ; \m^{(o)}_g,\s^{(o)}_g) + 
  \log\phi(\hat{\x}^{(m)}_{i,s}|\x^{(o)}_i ; \m^{(m)}_g,\s^{(m)}_g) 
\right\}, 
\label{MMCEM1:l}
\end{multline}
where $S$ is the MC sample size, $\hat{z}_{ig,s}$ is the simulated indicator variable, and $\hat{\x}^{(m)}_{i,s}$ is the imputed value. Comparing equation~\eqref{loglikComp::mixture} and equation~\eqref{MMCEM1:l}, it is clear that the latter is the complete-data log-likelihood of a GMM for the augmented dataset with dimension $(NS \times d)$.

Drawing $S$ indicator variables $ \hat{z}_{ig,s}$ for each observation $i$ from the conditional distribution $z_{ig}|\x_i^{(o)}$, the missing part is imputed $S$ times using the conditional distribution $\x_i^{(m)} |(\x_i^{(o)},\hat{z}_{ig,s})$, i.e. conditional on the simulated group membership $z_{ig,s}$ and on the observed part $\x_{i}^{(o)}$, to obtain $\hat{\x}^{(m)}_{i,s}$. We refer to this method as MMCEM1.

The second approach employs the MC approximation in a different way, together with the law of total expectation. For a general density function $h()$, the E-step of the MMCEM1 method described above is approximated as
\begin{equation}
\E[z_{ig}h(\x_i)] \approx \frac{1}{S}\sum_{s=1}^{S}z_{ig,s} h(\x_{i,s}).
\label{MMCEM1exp}
\end{equation}
Since equation~\eqref{MMCEM1exp} can be rewritten as
$$
\E[z_{ig} h(\x_i)]={\mathbb E} \bigg [z_{ig}  \mathbb E \left [ h(\x_i)|z_{ig}\right ] \bigg ],
$$
the inner expected value can be computed using MC approximations, and the outer expected value can be written in closed-form, i.e.
\begin{equation}
\E[z_{ig} h(\x_i)] 
\approx \E \left[ z_{ig} \frac{1}{S}\sum_{s=1}^{S}[h(\x_{i,s})|z_{ig}] \right]
= \hat{z}_{ig} \frac{1}{S} \sum_{s=1}^{S}[h(\x_{i,s})|z_{ig}].
\label{gen::MMCEM2}
\end{equation}
Using the approach in equation~\eqref{gen::MMCEM2}, the expected complete-data log-likelihood can be approximated as follows:
\begin{align}
Q(\h|\h^{t}) 
& = \sum_{i=1}^N \sum_{g=1}^G \E \left[ z_{ig} \left\{ \log\pi_g + \log  \phi(\x^{(o)}_i, \x^{(m)}_i ; \m_g,\s_g) \right\} \biggl| \x^{(o)} \right]\nonumber\\
& \approx \sum_{i=1}^N \sum_{g=1}^G \E \left[ \frac{z_{ig}}{S} \sum_{s=1}^S \left\{ \log\pi_g + \log\phi(\x^{(o)}_i,\hat{\x}^{(m)}_{i,sg} ; \m_g,\s_g \right\} \right] \nonumber\\
& = \frac{1}{S} \sum_{s =1}^S \sum_{n=1}^{N} \sum_{g=1}^G \hat{z}_{ig} \left\{ \log\pi_g + \log\phi(\x^{(o)}_i,\hat{\x}^{(m)}_{i,sg} ; \m_g, \s_g) \right\},
\label{MMCEM2:l}
\end{align}
where $\hat{z}_{ig}$ is the expected group membership indicator variable for observation $i$ computed using the conditional distribution of $z_{ig}|\x_i^{(o)}$, and $\hat{\x}^{(m)}_{i,sg}$ is the missing part of observation $i$ which is imputed $S$ times for each group $g=1\,\ldots,G$. Also in this case, equation~\eqref{MMCEM2:l} is the complete-data log-likelihood of the augmented dataset. We refer to this method as MMCEM2. 

Note that, since the standard M-step is unchanged, both approaches introduced in this Section allow to estimate the parameters for all the parsimonious GMMs obtained by adopting the different parameterisations of the covariance matrices. 
Next section provides further details concerning the proposed algorithms.

\section{MMCEM algorithms}
\label{sec:algo}

The approximations discussed in the previous Section yield two different methods for computing the E-step; see equation~\eqref{MMCEM1:l} and equation~\eqref{MMCEM2:l}. 
This section describes further computational details concerning the proposed extensions of the EM algorithm for GMMs with missing values.

\subsection{MMCEM1 algorithm}
\label{mMMCEM11:sec}

\subsubsection{Initialisation}
\label{init}

Initialisation is an important step in any iterative algorithm, especially when the function to maximise is multimodal, as in the case of GMMs. Proportions, means, and covariance matrices of the mixture components must be provided for starting the EM algorithm. They can be estimated from the fully observed dataset obtained after removing the missing observations. Thus, the classification matrix $\z$ is initialised using the observed part of each observation:
\begin{equation*}
\hat{z}_{ig}= \E \left[ z_{ig} | \x_i^{(o)} \right] = \frac{\pi_g \phi \left(\x_i^{(o)} ; \m^{(o)}_g, \s^{(o)}_g \right)}{\sum_{k=1}^G \pi_{k} \phi \left(\x_i^{(o)} ; \m^{(o)}_{k}, \s^{(o)}_{k} \right)}.
\end{equation*}

\subsubsection{E-step}

The imputation of the missing values is performed directly in the E-step using the following steps:
\begin{enumerate}
\item At iteration $t+1$, draw $(\hat{\z}_i^{1}, \ldots, \hat{\z}_i^{S})$  for each $i=1,\ldots,N$ from Multinomial distribution with parameter:
\begin{equation*}
\hat{z}_{ig,s} = \frac{\pi^t_g \phi \left(\x_i^{(o)} ; {\m^{(o)}_g}^t, {\s^{(o)}_g}^t \right)}{\sum_{k=1}^G \pi^t_{k} \phi \left(\x_i^{(o)} ; {\m^{(o)}_k}^t {\s^{(o)}_k}^t \right)},
\end{equation*}
where $\hat{z}_{ig,s}$ is the $s$th draw from the conditional probability of observation $i$ to belong to the cluster $g$, given the observed part of the data and the previous estimates of the parameters. Then, for observations with missing values only the observed part is considered.

\item After simulating the classification matrix, the missing values are imputed. Each missing values is simulated from $\phi(\x_i^{(m)} | \x_i^{(o)},\hat{z}_{ig,s}; \m_g, \s_g)$ for $s=1,2,\ldots,S$.

\item  The new augmented dataset $\hat{\x}_{(NS \times d)}$ is given by the union of the $S$ imputed datasets $\hat{\x}^s_{(N \times d)}$, and the new classification matrix $\hat{\z}_{(NS \times G)}$ is given by the union of the $S$ classification matrices $\hat{\z}^s_{(N \times G)}$ drawn at step 1. 
For instance, the generic observation $i$ containing at least one missing value can be represented as
\begin{equation*}
\begin{bmatrix}
x_i^{(o)} & x_{i,1}^{(m)} \\
x_i^{(o)} & x_{i,2}^{(m)} \\
\vdots & \vdots\\
x_i^{(o)} & x_{i,S}^{(m)}
\end{bmatrix},
\qquad
\begin{bmatrix}
\hat{z}_{i1,1}/S & \hat{z}_{i2,1}/S & \cdots & \hat{z}_{iG,1}/S\\
\hat{z}_{i1,2}/S & \hat{z}_{i2,2}/S & \cdots & \hat{z}_{iG,2}/S\\
\vdots & \vdots & \ddots & \vdots\\
\hat{z}_{i1,S}/S & \hat{z}_{i2,S}/S & \cdots & \hat{z}_{iG,S}/S\\
\end{bmatrix}.
\end{equation*}
Then, the new dataset $\hat{\x}_{(NS \times d)}$ and $\hat{\z}_{(NS \times G)}$ are used in the standard M-step.
\end{enumerate}

\subsubsection{M-step}
\label{mstep:alg}

The parameters and the classification matrix, obtained according to the MAP principle for assigning the observations to a given cluster, are computed using the standard M-step for GMMs using the augmented data $\hat{\x}_{(NS \times d)}$ and the classification matrix $\hat{\z}_{(NS \times G)}$.

\subsubsection{Iterations and convergence check}
\label{critConv}

Since standard convergence criteria for the EM algorithm cannot be used in this case because of MC error, iterations and convergence check are performed as described below:

\begin{enumerate}
\item EM steps are run for $T$ iterations, and only the parameters associated to the largest value of the log-likelihood are stored.
\item In the following EM steps, the parameters are updated when the log-likelihood improves compared to the previous step.
\item The algorithm is stopped if the log-likelihood does not increase for $K$ successive iterations.
\end{enumerate}

Two tuning parameters must be set in the above algorithm. If the initial number of ``warm-up'' iterations $T$ is large, the algorithm potentially explores a large part of the parameters space at the cost of increasing the computing time, and viceversa when $T$ is small. 
The second parameter $K$ specifies the number of ``stalled iterations'' allowed, so we can control how stringent is the adopted criterion for declaring the convergence of the algorithm. A sensible choice of these tuning parameters is needed to balance the efficiency and accuracy of the MCEM.
 
\subsection{MMCEM2 algorithm}
\label{mMMCEM2:sec}

\subsubsection{Initialisation, convergence and M-step}

MMCEM1 and MMCEM2 differ only for the E-step, then both methods share the same initialisation, iterations, convergence criterion, and the M-step (see, respectively, Section \ref{init}, \ref{critConv}, and \ref{mstep:alg}). In the following we provide details only for the E-step.

\subsubsection{E-step}

As in the MMCEM1 algorithm, the imputation is performed during the E-step. 
The algorithm used to build the augmented dataset is the following:
\begin{enumerate}

\item For each observation, simulate $S$ values $x_{ig,s}$ from
\begin{equation*}	
\x_i^{(m)} | (\x_i^{(o)}, z_{ig}=1 ; {\m^{(0)}_g}^t,{\s^{(o)}_g}^t),
\end{equation*}
for $g=1, \ldots, G$.

\item Construct an augmented data set where each observation is represented as follows:
\begin{equation*}
\left[\begin{array}{cc}
x_i^{(o)} & x_{i,11}^{(m)} \\
x_i^{(o)} & x_{i,12}^{(m)} \\
\vdots & \vdots\\
x_i^{(o)} & x_{i,1S}^{(m)} \\[0.5ex] 
\hdashline\\[-2ex] 
x_i^{(o)} & x_{i,21}^{(m)} \\
x_i^{(o)} & x_{i,22}^{(m)} \\
\vdots & \vdots\\
x_i^{(o)} & x_{i,2S}^{(m)} \\[0.5ex] 
\hdashline\\[-2ex] 
\vdots & \vdots \\[0.5ex] 
\hdashline\\[-2ex] 
x_i^{(o)} & x_{i,G1}^{(m)} \\
x_i^{(o)} & x_{i,G2}^{(m)} \\
\vdots & \vdots\\
x_i^{(o)} & x_{i,GS}^{(m)} \\
\end{array}\right]
\qquad
\left[\begin{array}{cccc}
\hat{z}_{i1}/S & 0 & \cdots & 0\\
\hat{z}_{i1}/S & 0 & \cdots & 0\\
\vdots & \vdots & \ddots & \vdots\\
\hat{z}_{i1}/S & 0 & \cdots & 0 \\[0.5ex] 
\hdashline\\[-2ex] 
0 & \hat{z}_{i2}/S & \cdots & 0\\
0 & \hat{z}_{i2}/S & \cdots & 0\\
\vdots & \vdots & \ddots & \vdots\\
0 & \hat{z}_{i2}/S & \cdots & 0 \\[0.5ex] 
\hdashline\\[-2ex] 
\vdots & \vdots & \ddots & \vdots \\[0.5ex] 
\hdashline\\[-2ex] 
0 & 0 & \cdots & \hat{z}_{iG}/S\\
0 & 0 & \cdots & \hat{z}_{iG}/S\\
\vdots & \vdots & \ddots & \vdots\\
0 & 0 & \cdots & \hat{z}_{iG}/S\\
\end{array}\right]
\end{equation*}
\end{enumerate}

\section{Examples}

In this Section, the proposed algorithms are evaluated on both simulated and real datasets to assess their performance in terms of clustering and density estimation. 

The software package \pkg{mclust}, freely available on CRAN (\url{https://cran.r-project.org/web/packages/mclust/index.html}) for the R language \citep{RStat}, provides fitting of finite mixture of Gaussian distributions through the use of the EM algorithm \citep{mclust2}. In particular, the function \verb|mstep()| can be used to perform the maximisation step for each of the fourteen models in the parsimonious GMM family generated by the eigen-decomposition of the covariance matrices discussed in Section~\ref{sec:gmm}. This function, together with our code that implements the two versions of the E-step, have been used to build the algorithms described in Section~\ref{sec:algo}.  

The methods included in the comparison are:
\begin{enumerate}
	\item EM algorithm with MC approximations of the E-step as presented in Section \ref{mMMCEM11:sec} (MMCEM1).
	\item EM algorithm with MC approximations of the E-step as presented in Section \ref{mMMCEM2:sec} (MMCEM2).
	\item Multiple imputation \citep{schafer1997analysis}, where $N_{imp}$ different imputed datasets are generated, and for each of these the standard EM algorithm is applied to estimate the density and the final clustering (GMMMI). The \verb|imputeData()| function in the \pkg{mclust} package is used to impute the missing values, and the \verb|Mclust()| function is used to estimate GMMs on the imputed dataset. A label switching strategy is also implemented using the majority vote to assign the observations to the different clusters. The number of multiple imputations is set at $50$.
	\item Gaussian mixture on the original dataset before introducing the missing values (GMM). The \verb|Mclust()| function from \pkg{mclust} package is used to estimate the parameters of the Gaussian mixture. These estimates represent the benchmark to which the different methods should tend in the presence of missing values; the closer the values are to these estimates, the better a method can recover the missing information.
\end{enumerate}

MC sample size is one of the most important tuning parameter in our proposed algorithms. This parameter must balance the precision of the method and the computational efficiency. Large MC sample sizes imply a higher probability of convergence to the true value, and therefore greater precision. In contrast, small MC sample sizes improve the speed of the algorithm to the detriment of the accuracy of the estimates.
In our experiments we set the MC sample size to $S = 10$. This relatively small value provides a conservative assessment of the precision and efficiency of the MMCEM1 and MMCEM2 algorithms. Moreover, the ``warm-up'' parameter is set at $T=10$, and the ``stalled iterations'' parameter is set at $K=1$. Since the largest improvements of log-likelihood are likely to happen in the initial iterations of the EM algorithm, we tried to replicate conditions similar to the standard EM algorithm. Clearly, larger MC sample sizes and larger values of the tuning parameters could only improve model fitting, at the cost of increasing the computing effort.

\subsection{Synthetic data}
\label{sec:synthetic}

Simulated datasets are generated from eight different scenarios using a mixture of Gaussian distributions with number of variables $d = 2$ and number of mixture components $G = 3$. Details for each scenario are the following:

\begin{enumerate}[label= (\alph*)]

\item Three well-separated clusters from a Gaussian mixture with mean vector for each component given by $\boldsymbol{\mu}_{1} = [4,4]\T$, $\boldsymbol{\mu}_{2} = [0,-4]\T$, $\boldsymbol{\mu}_{3} = [-4,4]\T$, and common covariance matrix: 
	\begin{equation*}
	\s =
	\begin{bmatrix}
	1 & 0.25 \\
	0.25 & 1 
	\end{bmatrix}.
	\end{equation*}
	This correspond to model EEE in the \pkg{mclust} nomenclature. The mixing probabilities are all equal to $1/3$.

\item Three well-separated clusters with strongly unbalanced mixture of Gaussians with prior probabilities set to $\boldsymbol{\pi} = (0.7,0.25,0.05)$, and with the remaining parameters set as in the previous scenario.

\item Three-groups case with two overlapping clusters having mean vector for each component given by $\boldsymbol{\mu}_{1} = [2,3.5]\T$, $\boldsymbol{\mu}_{2} = [-2,3]\T$, and $\boldsymbol{\mu}_{3} = [0,2]\T$, and common covariance matrix: 
	\begin{equation*}
	\s =
	\begin{bmatrix}
	1 & -0.5 \\
	-0.5 & 0.4 
	\end{bmatrix}.
	\end{equation*}
	This corresponds to model EEE in the \pkg{mclust} nomenclature. The mixing probabilities are all equal to $\frac{1}{3}$.
	
\item Three-groups case with two overlapping clusters with strongly unbalanced clusters with prior probabilities set to $\boldsymbol{\pi} = (0.7,0.25,0.05)$, and with the remaining parameters set as in the previous scenario.	

\item Three well-separated clusters with mean for each component given by $\boldsymbol{\mu}_{1} = [4,4]\T$, $\boldsymbol{\mu}_{2} = [0,-4]\T$, $\boldsymbol{\mu}_{3} = [-4,4]\T$, and unconstrained covariance matrices:
	\begin{equation*}
	\s_1 =
	\begin{bmatrix}
	1 & 0 \\
	0 & 1 
	\end{bmatrix},
    \qquad
	\s_2 =
	\begin{bmatrix}
	1.5 & -1 \\
	-1 & 2 
	\end{bmatrix},
	\quad
	\s_3 =
	\begin{bmatrix}
	1 & 0.5 \\
	0.5 & 3 
	\end{bmatrix}.
	\end{equation*}
	This corresponds to model VVV in the \pkg{mclust} nomenclature. The mixing probabilities are all equal to $1/3$.
	
\item Three well-separated clusters with strongly unbalanced mixture of Gaussians with prior probabilities set to $\boldsymbol{\pi} = (0.7,0.25,0.05)$, and with the remaining parameters set as in the previous scenario.
	
\item Three-groups case with two overlapping clusters having mean vector for each component given by $\boldsymbol{\mu}_{1} = [1.5,2]\T$, $\boldsymbol{\mu}_{2} = [-3,0]\T$, $\boldsymbol{\mu}_{3} = [-4,4]\T$, and unconstrained covariance matrices:
	\begin{equation*}
	\s_1 =
	\begin{bmatrix}
	1 & 0 \\
	0 & 1 
	\end{bmatrix},
	\quad
	\s_2 =
	\begin{bmatrix}
	1 & -1 \\
	-1 & 1.5 
	\end{bmatrix},
	\quad
	\s_3 =
	\begin{bmatrix}
	1 & 0.5 \\
	0.5 & 3 
	\end{bmatrix}.
	\end{equation*}
	This corresponds to model VVV in the \pkg{mclust} nomenclature. The mixing probabilities are all equal to $1/3$.

\item Three-groups case with two overlapping clusters with strongly unbalanced clusters and prior probabilities set to $\boldsymbol{\pi} = (0.7,0.25,0.05)$, and with the remaining parameters set as in the previous scenario.
\end{enumerate}

Scenarios (a)-(b)-(e)-(f) are relatively simpler cases compared to scenarios (c)-(d)-(g)-(h). In the latter cases, because clusters have substantial overlap, detecting the exact number of components can be particularly difficult.
The unbalanced cases in scenarios (b)-(d)-(f)-(h) are used to assess the effect of missing values when some clusters have small sizes. 

\begin{figure}[htb]
\centering
\includegraphics[width=\textwidth]{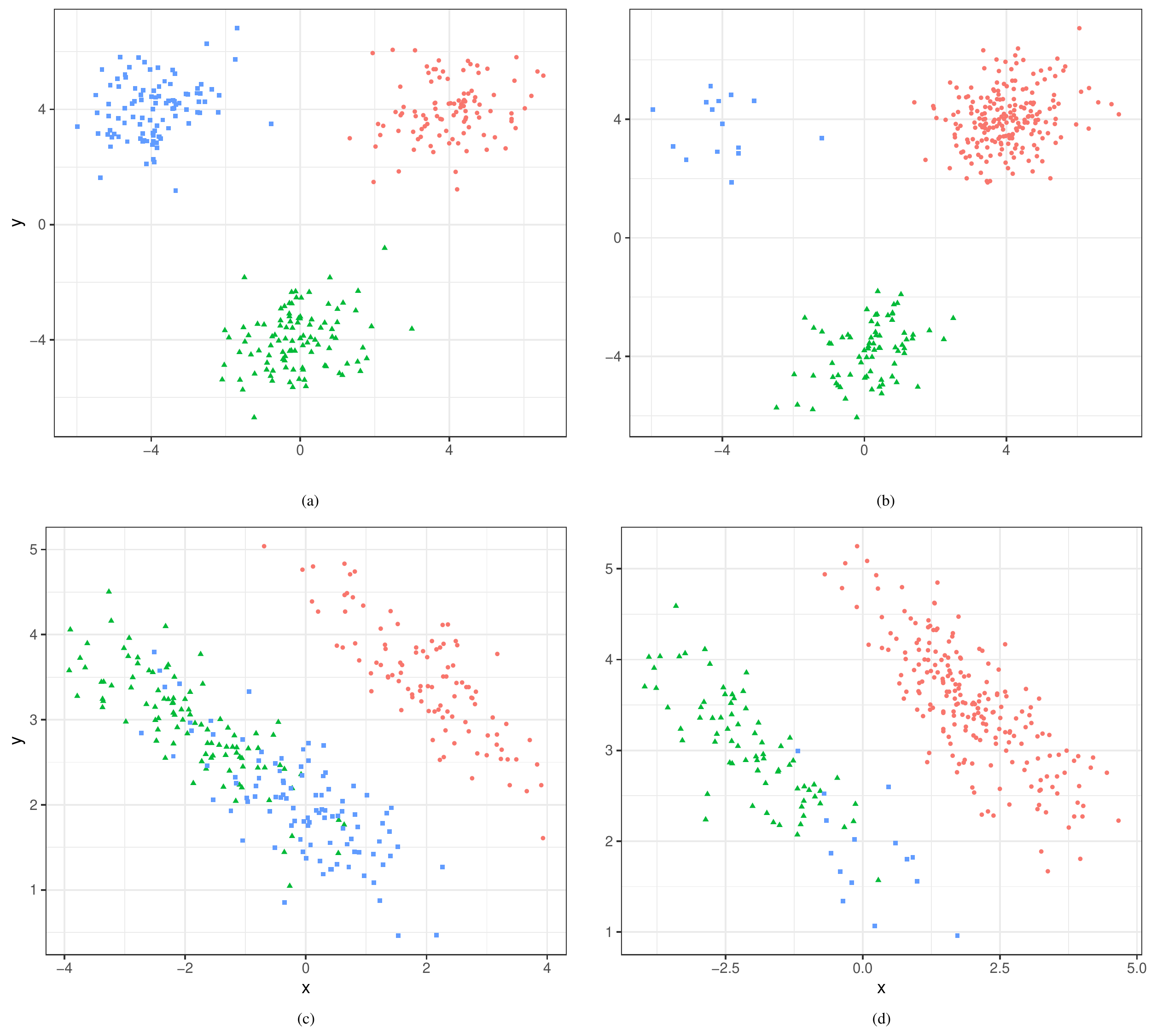}
\caption{Some simulated datasets from scenarios having EEE configurations: (a) balanced clusters; (b) unbalanced clusters; (c) balanced overlapping clusters; (d) unbalanced overlapping clusters.}
\label{fig:simEEE}
\end{figure}

\begin{figure}[htb]
\centering
\includegraphics[width=\textwidth]{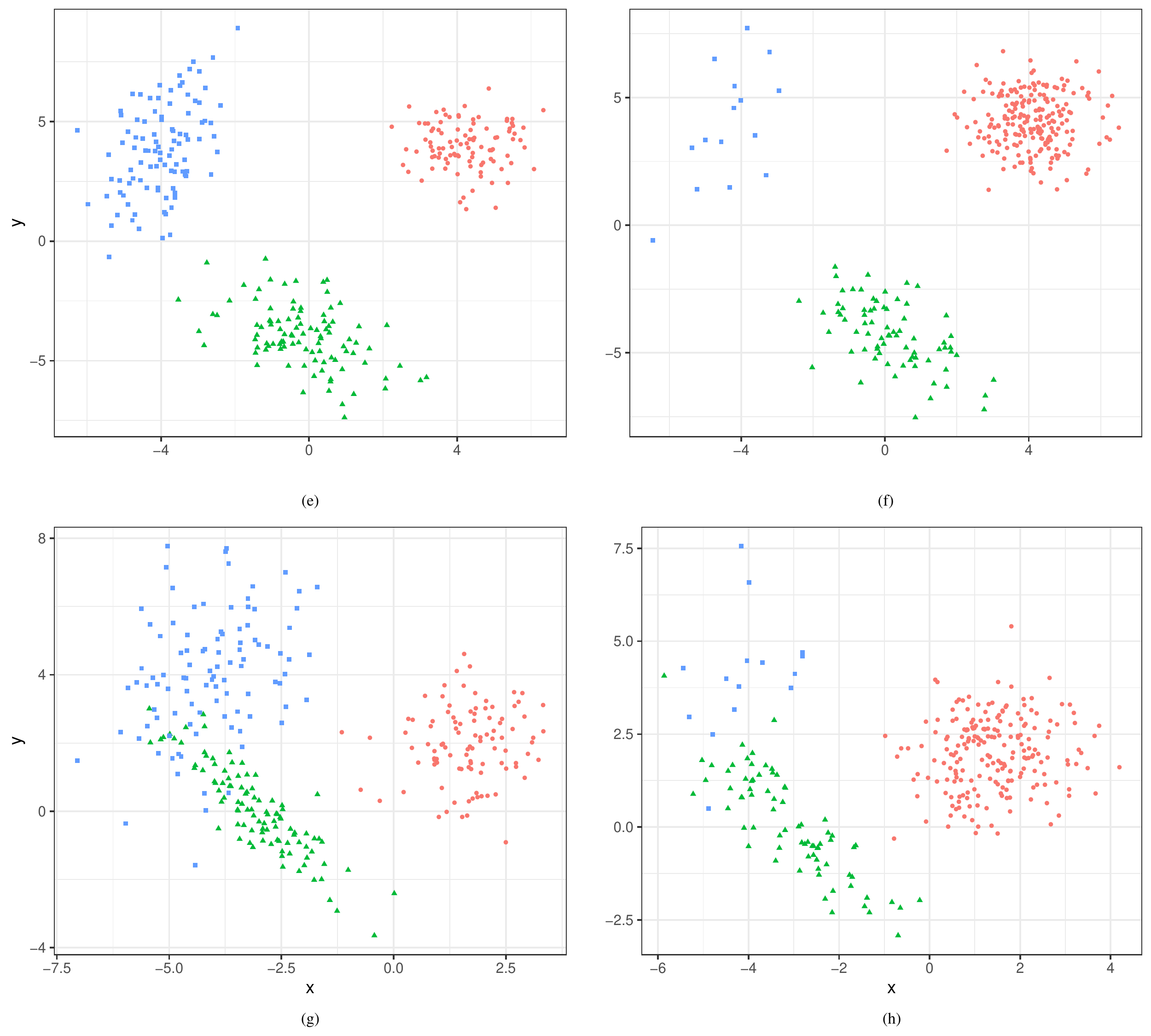}
\caption{Some simulated datasets from scenarios having VVV configurations: (e) balanced clusters; (f) unbalanced clusters; (g) balanced overlapping clusters; (h) unbalanced overlapping clusters.}
\label{fig:simVVV}
\end{figure}

For all the above scenarios the number of observations is set to $N=300$, and each scenario was replicated $1000$ times. Figures \ref{fig:simEEE}-\ref{fig:simVVV} show some examples of simulated datasets. 
 
A missing mechanism is applied to each dataset with missing percentage set to about $50\%$, i.e. approximately half of the observations have at least a missing value. Details differ depending on whether MCAR or MAR mechanism is used. 

For the MCAR mechanism, incomplete observations are randomly selected, and for each observation the variable to contain the missing value is also selected at random. This two steps guarantee that the mechanism is MCAR because the missing values are a random sample of all data values.

To generate data under the MAR mechanism the following process is adopted. 
Let $\boldsymbol{M}$ be a $(N \times d)$ matrix of indicator variables, such that $m_{ij}=0$ if $x_{ij}$ is missing, and 1 otherwise. 
From the definition in \citet{schafer1997analysis}, the missing values are supposed to be MAR if:
\begin{equation}
P(\boldsymbol{M}| \x^{(o)},\x^{(m)}; \h) = P(\boldsymbol{M} | \x^{(o)}; \h).
\label{MAR_prob}
\end{equation} 
However, to replicate a MAR mechanism it is necessary to estimate the probabilities in \eqref{MAR_prob}. Since we are generating two-dimensional datasets, without loss of generality, we may assume that the first variable is completely observed, and the second variable contains all the missing values. 
The probability of a missing value on the second variable for the $i$th observation can be modelled using an inverse \textit{logit} transformation:
\begin{equation*}
P(m_{i2} = 0 | \x^{(o)}; \h) = \frac{\exp(\beta x_{i1})}{1 + \exp(\beta x_{i1})},
\end{equation*}
for all $i = 1,\ldots,N$. The value $\beta = 0.01$ guarantees that on average about 50\% of the observations have a missing value. Then, these probabilities are used to randomly select from a Bernoulli distribution those observations that have a missing value in the second variable. 

The performance of the methods under investigation is evaluated in terms of both density estimation and clustering accuracy.
To assess the accuracy of density estimates the Kullback-Leibler divergence \citep[KL;][]{kullback1951information} is used. 
This is a dissimilarity measure between two probability density functions.
Let $f(\x)$ and $g(\x)$ be two density functions, then the KL divergence is defined as follows:
\begin{equation}
D(f||g) = \int f(\x)\log\frac{f(\x)}{g(\x)}d\x,
\label{KL:missing}
\end{equation}
where $D(f||g) = 0$ if the two densities are equal. In general, $D(f||g) > 0$ and gets larger as the diversity between the two densities increases.
The density $f(\x)$ is taken to be the true density of the simulated data, whereas $g(\x)$ is the density estimated using one of the methods under comparison. Then, to have a good density estimation method, the KL should be as much as possible close to zero. Since GMMs do not have a closed-form expression for \eqref{KL:missing}, a MC approximation must be used \citep{hershey2007approximating}. 

To compare the estimated classification with the true clusters, the Adjusted Rand Index \citep[ARI;][]{hubert1985comparing} is used. This measures the agreement between two partitions, with expected value 0 in case of random partitions, and a value equal to 1 in case of a perfect agreement. Thus, the true partition is compared with the classification provided by the GMM methods under comparison in the presence of missing values.

Models are estimated either by fixing the number of components and the parameterisation of component-covariances used to generate the data, and then by selecting the number of mixture components by the the Bayesian Information Criterion \citep[BIC;][]{schwarz1978estimating}. A last configuration is considered when both the number of components and the parsimonious decomposition of component-covariance matrix are selected by BIC.
The last two situations allow to investigate the influence of missing values on GMMs parameters estimation when either the number of groups is not known a priori, or the component-covariance matrix, or both.

\begin{figure}[htb]
\centering
\includegraphics[width=\textwidth]{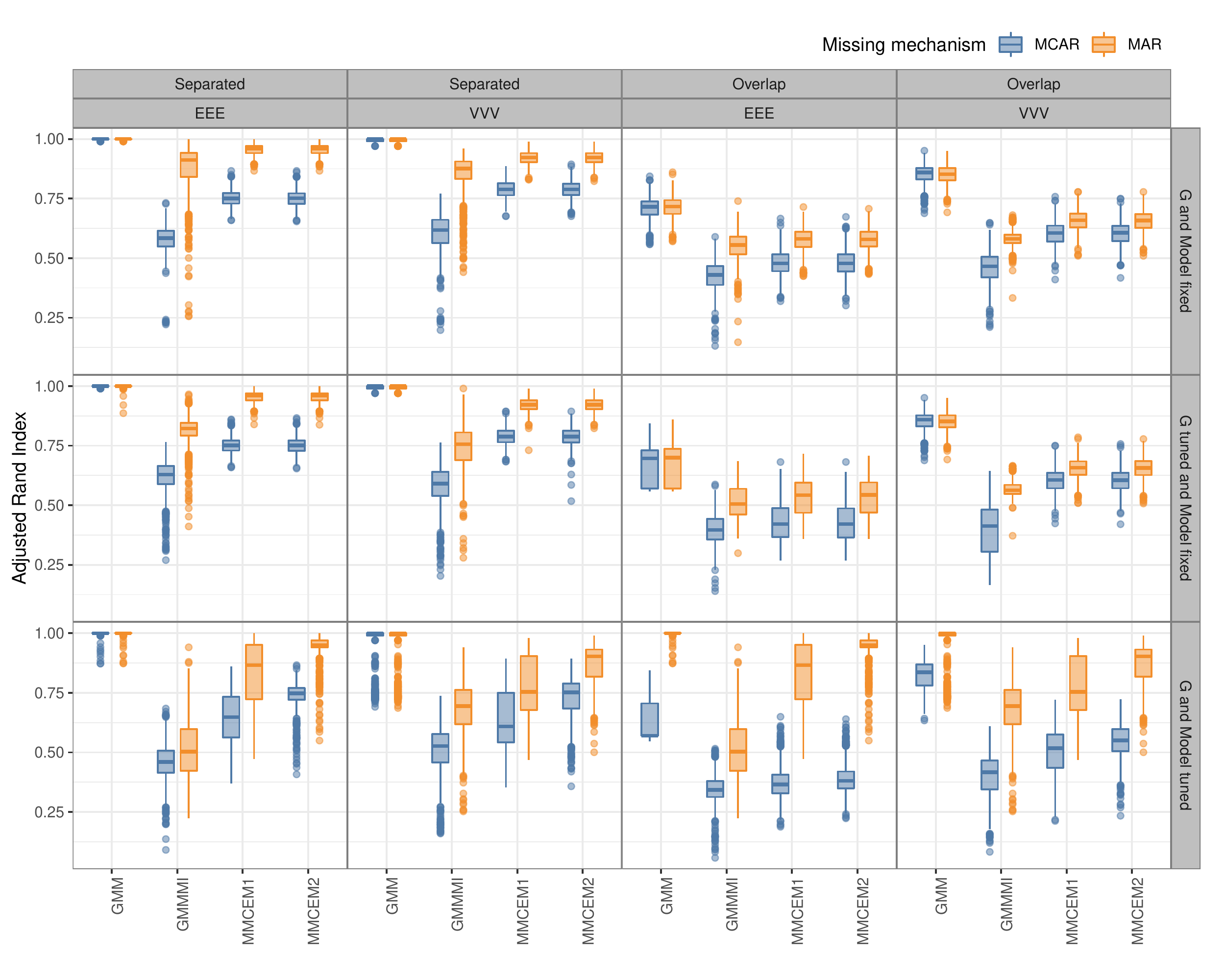}
\caption{Box-plots of ARI values from the simulation study for the 3-cluster cases with balanced mixture proportions under different missing mechanisms (larger values are better).}
\label{ARI_BALANCED}
\end{figure}

\begin{figure}[htb]
\centering
\includegraphics[width=\textwidth]{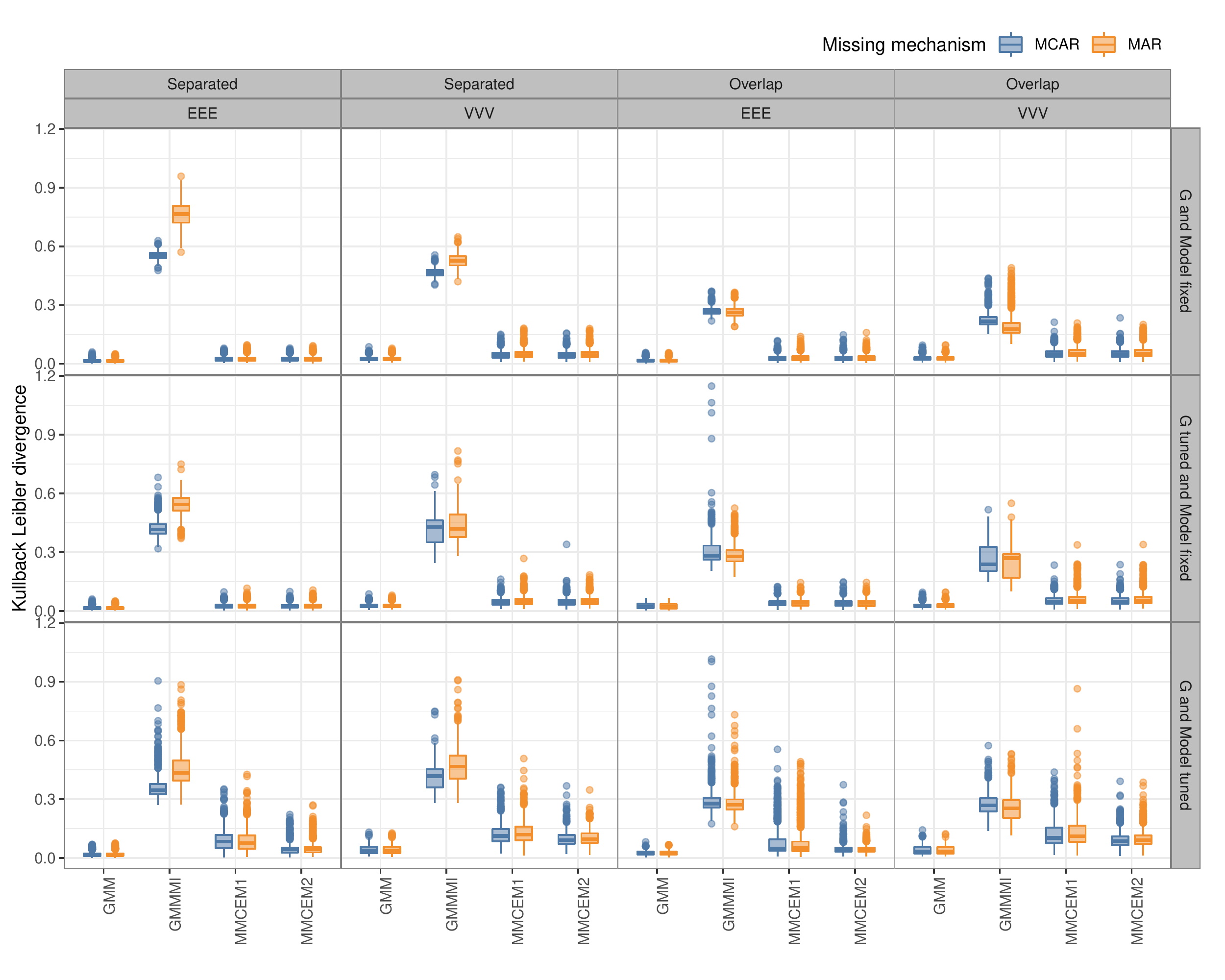}
\caption{Box-plots of KL values from the simulation study for the 3-cluster case with balanced mixture proportions under different missing mechanisms (smaller values are better).}
\label{KL_BALANCED}
\end{figure}

\begin{figure}[htb]
\centering
\includegraphics[width=\textwidth]{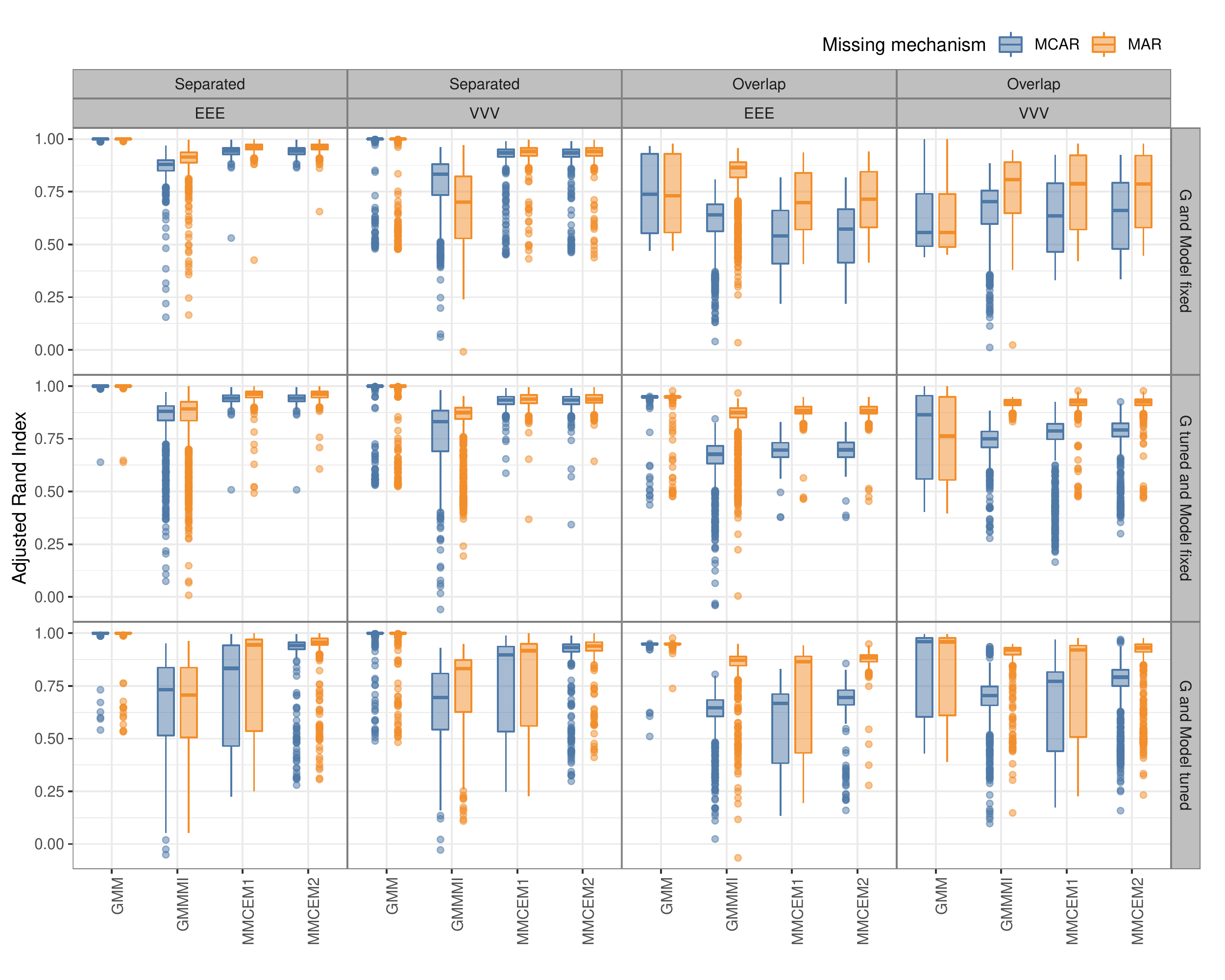}
\caption{Box-plots of ARI values from the simulation study for the 3-cluster case with unbalanced mixture proportions under different missing mechanisms (larger values are better).}
\label{ARI_UNBALANCED}
\end{figure}

\begin{figure}[htb]
\centering
\includegraphics[width=\textwidth]{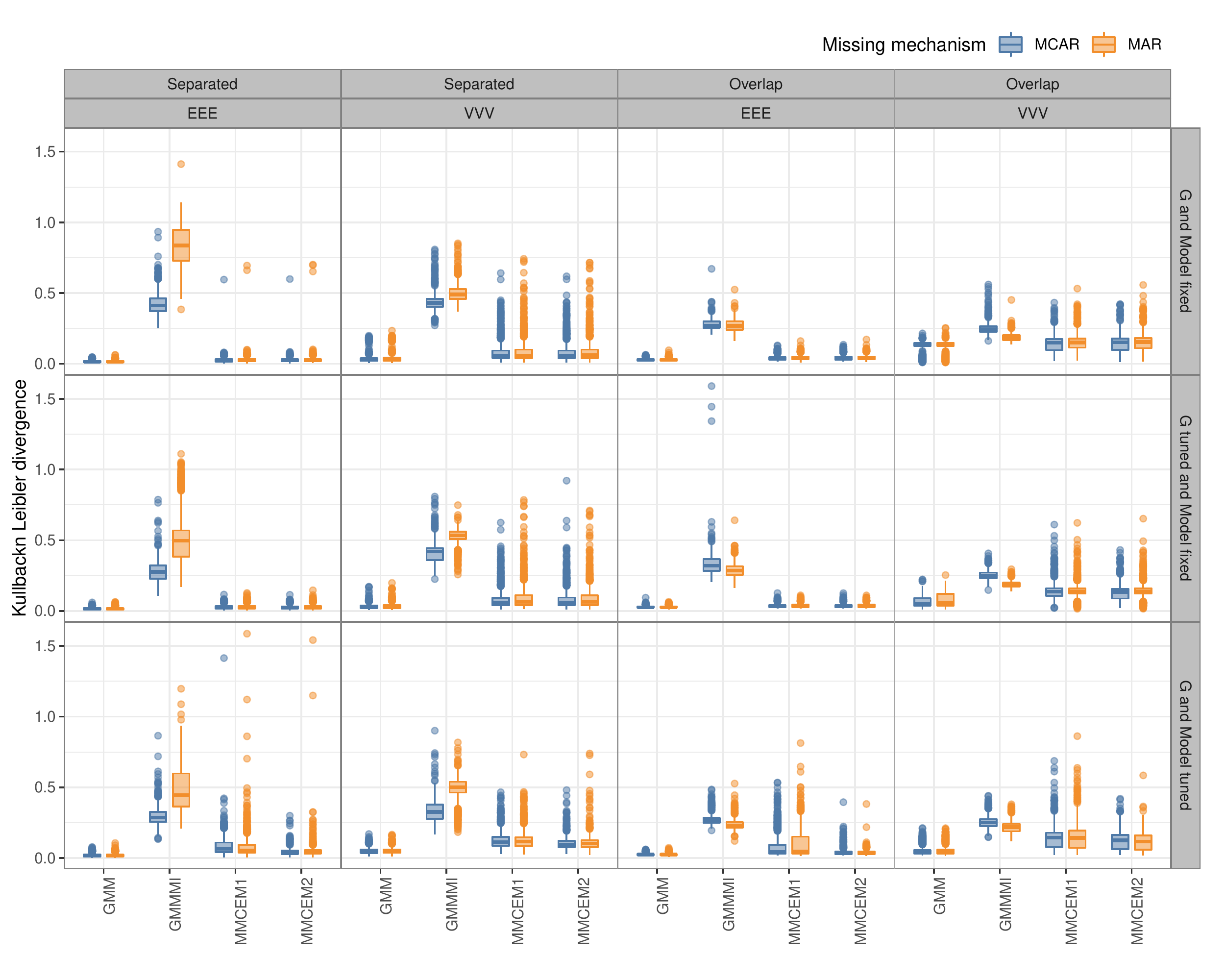}
\caption{Box-plots of KL values from the simulation study for the 3-cluster case with unbalanced mixture proportions under different missing mechanisms (smaller values are better).}
\label{KL_UNBALANCED}
\end{figure}

Figures \ref{ARI_BALANCED}--\ref{KL_UNBALANCED} show the results of the simulation study using box-plots for each method in each scenario. 

In general, the proposed methods clearly outperform the multiple imputation approach in all scenarios in terms of density estimate accuracy. To this goal, MMCEM1 and MMCEM2 are essentially equivalent and close to the estimates obtained using the complete dataset (GMM).
The same also applies to cluster identification, with few exceptions where the three methods are roughly comparable. Multiple imputation (GMMMI) appears to be no worse than MMCEM1 and MMCEM2 only when clusters are unbalanced and overlapping.

\begin{figure}[htb]
\centering
\includegraphics[width=\textwidth]{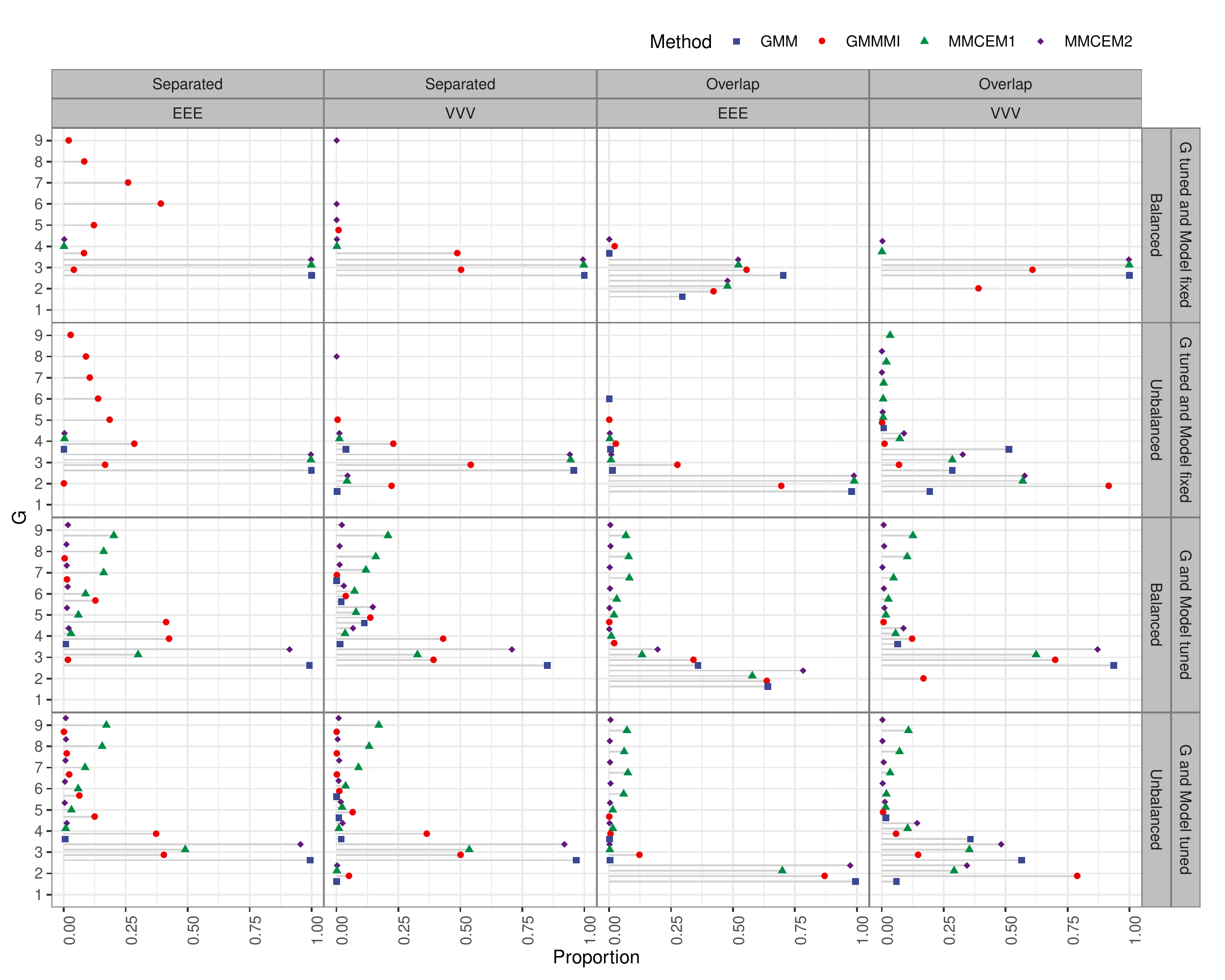}
\caption{Distribution of the number of mixture components selected by BIC under the MCAR missing mechanisms.}
\label{fig:distribution_G_MCAR}
\end{figure}

\begin{figure}[htb]
\centering
\includegraphics[width=\textwidth]{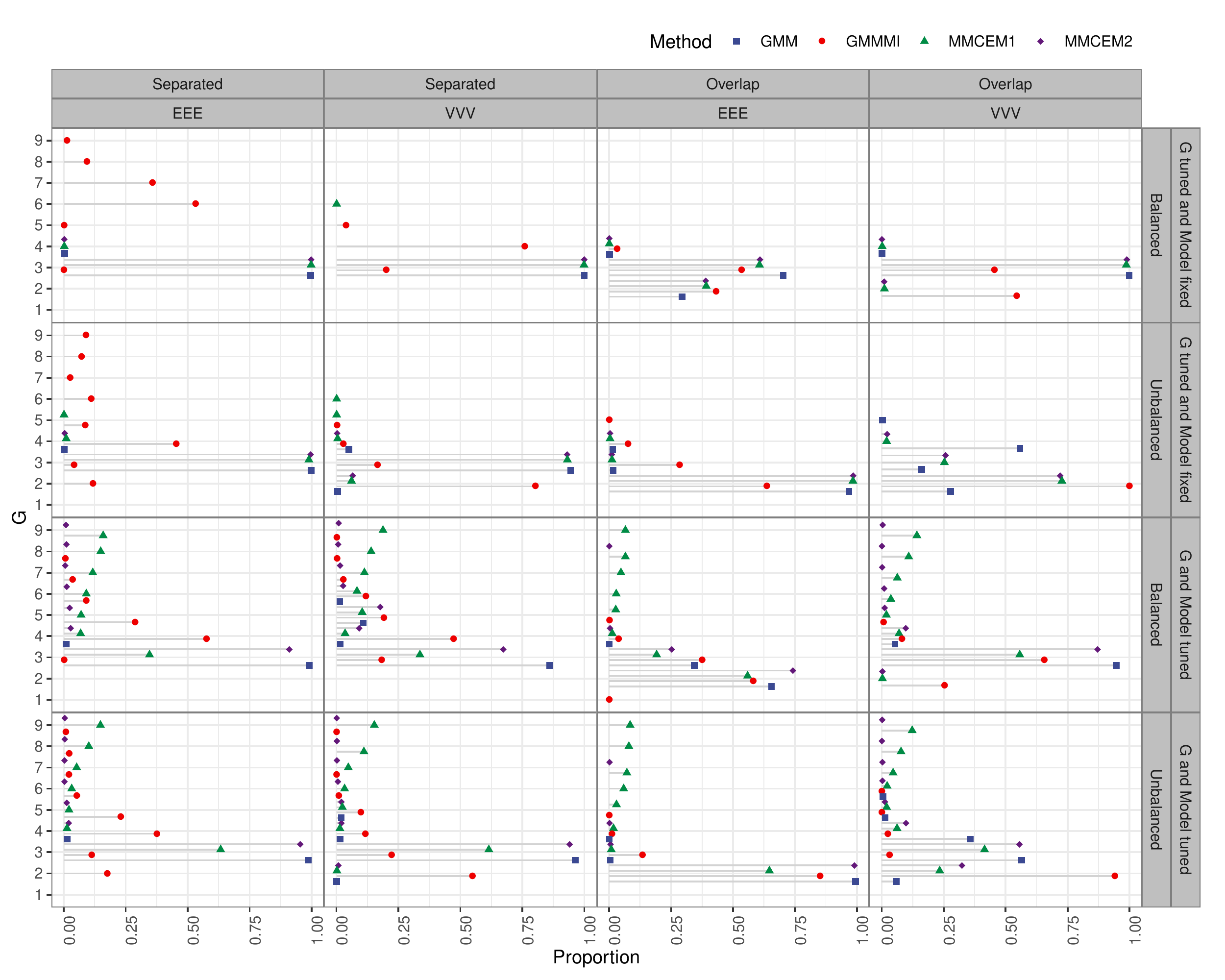}
\caption{Distribution of the number of mixture components selected by BIC under the MAR missing mechanisms.}
\label{fig:distribution_G_MAR}
\end{figure}

When the number of groups is selected by BIC, the proposed methods again perform better than the multiple imputation approach, both in terms of cluster accuracy and density estimation. In particular, by removing the number of clusters the MMCEM methods outperform in term of classification the multiple imputation also in the overlapping case with unbalanced clusters. 
In addition, as expected, GMM selects three groups, whereas MMCEM1 and MMCEM2 select three components in the majority of cases. Conversely, the multiple imputation approach (GMMMI) tends to select more components that the other methods in the well separated clusters, and less components in the overlapping clusters (see Figures \ref{fig:distribution_G_MCAR}--\ref{fig:distribution_G_MAR}).

If the covariance matrix is constrained to be equal among the components, the multiple imputation approach tends to select more clusters than the original groups. This may be due to the imputation step that generates points that fill the gaps between the clusters. As a consequence, imposing the ellipsoids to be equal increases the number of mixture components required. 

When both the number of components and the covariance model are unknown and selected by the BIC criterion, our MMCEM methods tend to outperform the multiple imputation, with values close to the estimates provided by the GMM on the original data. This suggests that our methods seem to be able to recover the original structure of the data.

Another element arises from the simulations. The MMCEM2 algorithm appears to perform better than the MMCEM1 algorithm, with smaller standard errors, indicating a more precise method. Such behaviour can be noted also in the distribution of the number of estimated components in Figures \ref{fig:distribution_G_MCAR}--\ref{fig:distribution_G_MAR}; in most cases MMCEM2 selects the correct number of clusters, whereas MMCEM1 has much more variability in selecting the number of components.

\clearpage

\subsection{Diabetes data} 
\label{sec:diabetes}

\citet{reaven1979attempt} provided data from a diabetes study conducted at the Stanford Clinical Research Center. Blood chemistry measurements were recorded on 145 non-obese adult subjects, namely the area under plasma glucose curve, the area under plasma insulin curve, and the steady state plasma glucose level. 
After further analysis, the patients were classified into three groups: a group of patients suffering from overt diabetes (Overt), a group affected by chemical diabetes (Chemical), and a last group made of patients without diabetes (Normal). The dataset is available in the R package \pkg{mclust}.

Missing values are introduced according to the MCAR mechanism under two different missing data patterns. 
In the first data pattern scenario, a single missing value is randomly assigned to a given percentage of sample observations. By setting this percentage at approximately $30\%$ and $50\%$, we get, respectively, $43$ and $72$ observations with a single missing value out of $145$ observations.
In the second data pattern scenario, the percentage of missing values refer to the overall number of measurements. Hence, setting the percentage to approximately $30\%$ and $50\%$, the total number of missing values randomly entered into the data matrix are $130$ and $217$, respectively, out of $435$ total measurements.

In this real data analysis example, we aim at comparing the clustering performance of the GMM fitted on the original data, as in the simulation studies, against the proposed methods, i.e. MCEM1 and MCEM2, the multiple imputation approach, and the GMM fitted on the data obtained after removing the observations with at least one missing value (GMMD). The performance of these methods are evaluated using the ARI. Furthermore, the BIC criterion is employed to select both the number of clusters and the parsimonious within-component covariance matrices.

\begin{figure}[htb]
\centering
\includegraphics[width=0.9\textwidth]{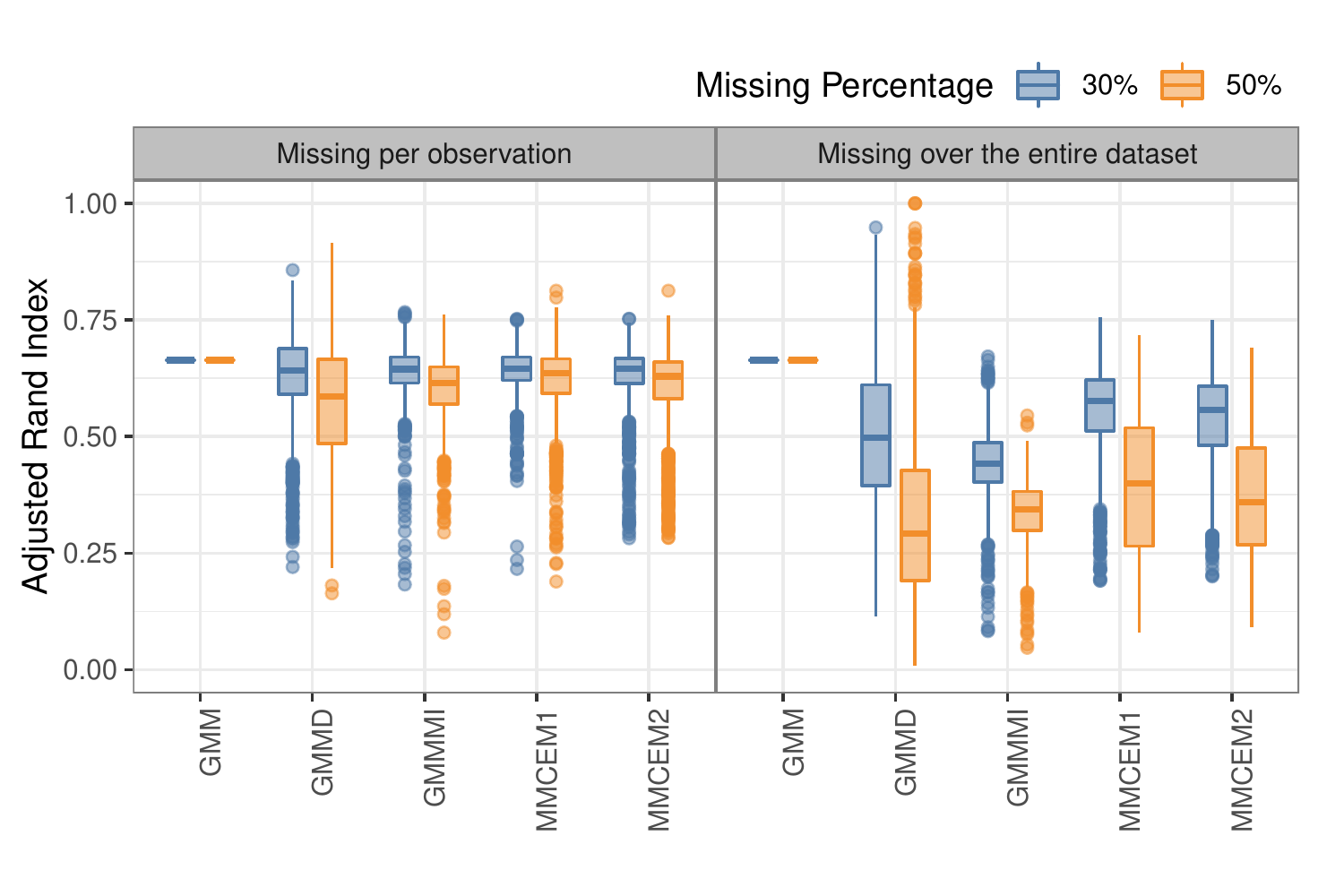}
\caption{Box-plots of ARI values for the Diabetes data with different MCAR mechanism scenarios and missing percentage (larger values are better). Both the number of mixture components and the component-covariance model are selected by BIC.}
\label{ARI_DIABETES}
\end{figure}

Figure~\ref{ARI_DIABETES} shows the results for the Diabetes data after applying the missing procedure $1,000$ times outlined above. 
In the first missing data scenario, where missing values are randomly assigned to observations, so each row of the data matrix has at most one missing value, the methods perform in a similar way, and they are pretty close to the case of GMM estimated on the full dataset (see left panel of Figure~\ref{ARI_DIABETES}).
In the second scenario, where the percentage of missing values is distributed over the entire dataset, the proposed MMCEM methods appear to outperform the other methods based on listwise-deletion or multiple imputation (see right panel of Figure~\ref{ARI_DIABETES}).

\section{Discussion}

In this paper we proposed two different algorithms to estimate GMMs in the presence of missing values by exploiting the well-known EM algorithm. 
Both algorithms employ Monte Carlo methods during the E-step to build augmented datasets via stochastic missing values imputation. These are then used in the standard M-step, thus allowing to obtain parameters estimates for any  parsimonious component-covariance matrix structure available for Gaussian mixtures. 

In general, the proposed methods seem to outperform the multiple imputation approach, both in terms of clustering and density estimation. 
The MMCEM1 and MMCEM2 algorithms are basically equivalent when the number of mixture components and the covariance structure are known. 
When these are unknown and, consequently, are selected by BIC, the MMCEM2 procedure provides more accurate clustering partitions and density estimates. 

On the other hand, the proposed algorithms are highly demanding in terms of computational cost. For high-dimensional dataset the procedures could need a large number of iterations during the model estimation phase, because of data augmentation in the imputation step that strongly depends on the number of observations and the sample size of the MC approximation.
For this reason a more efficient method could be investigated. In addition, since the proposed methods are based on the MAR and MCAR assumptions, another future development might consider MNCAR missing mechanism. 
 
\bibliography{Handling_missing_data_in_model-based_clustering}

\end{document}